%% file: main.tex
\documentclass{article}
\usepackage{ijcai25}

\usepackage{times}
\usepackage{soul}
\usepackage{url}
\usepackage[hidelinks]{hyperref}
\usepackage[utf8]{inputenc}
\usepackage[small]{caption}
\usepackage{graphicx}
\usepackage{amsmath}
\usepackage{amsthm}
\usepackage{booktabs}
\usepackage{algorithm}
\usepackage{algorithmic}
\usepackage[switch]{lineno}
\usepackage{xcolor}

\usepackage{multirow}
\usepackage{multicol}
\usepackage{makecell}
\usepackage{threeparttable}
\usepackage{pifont}
\usepackage{enumitem}
\usepackage{xcolor}
\definecolor{iceblue}{RGB}{173, 216, 230}

\title{A Survey on Computational Pathology Foundation Models: Datasets, Adaptation Strategies, and Evaluation Tasks}

\author{
    Dong Li$^1$\thanks{Equal contribution.}, 
    Guihong Wan$^{2,3*}$, 
    Xintao Wu$^4$,
    Xinyu Wu$^1$,  
    Ajit J. Nirmal$^{3,5}$, 
    Christine G. Lian$^{3,6}$, \\
    Peter K. Sorger$^3$, 
    Yevgeniy R. Semenov$^{2,3}$\thanks{Corresponding authors.}, 
    Chen Zhao$^{1\dagger}$
    \affiliations
    $^1$Department of Computer Science, Baylor University, TX, USA\\
    $^2$Department of Dermatology, Massachusetts General Hospital, Harvard Medical School, MA, USA\\
    $^3$Laboratory of Systems Pharmacology, Harvard Medical School, MA, USA\\
    $^4$Department of Electrical Engineering and Computer Science, University of Arkansas, AR, USA\\
    $^5$Department of Dermatology, Brigham and Women's Hospital, Harvard Medical School, MA, USA\\
    $^6$Department of Pathology, Brigham and Women's Hospital, Harvard Medical School, MA, USA
    \emails
    \{dong\_li1, chen\_zhao\}@baylor.edu,
    \{gwan, ysemenov\}@mgh.harvard.edu,
    xintaowu@uark.edu
}

\begin{document}

\maketitle

\begin{abstract}
    Computational pathology foundation models (CPathFMs) have emerged as a powerful approach for analyzing histopathological data, leveraging self-supervised learning to extract robust feature representations from unlabeled whole-slide images. These models, categorized into uni-modal and multi-modal frameworks, have demonstrated promise in automating complex pathology tasks such as segmentation, classification, and biomarker discovery. However, the development of CPathFMs presents significant challenges, such as limited data accessibility, high variability across datasets, the necessity for domain-specific adaptation, and the lack of standardized evaluation benchmarks. This survey provides a comprehensive review of CPathFMs in computational pathology, focusing on datasets, adaptation strategies, and evaluation tasks. We analyze key techniques, such as contrastive learning, masked image modeling and multi-modal integration, and highlight existing gaps in current research. Finally, we explore future directions from four perspectives for advancing CPathFMs. This survey serves as a valuable resource for researchers, clinicians, and AI practitioners, guiding the advancement of CPathFMs toward robust and clinically applicable AI-driven pathology solutions.
\end{abstract}

\section{Introduction}
\label{sec:introduction}
\input{sections/introduction.tex}

\section{Background}
\label{sec:background}
\input{sections/background.tex}

\vspace{-3mm}
\section{Pre-training Datasets in CPathFMs}
\label{sec:datasets}

\input{sections/pretraining_datasets}

\vspace{-3mm}
\section{Adaptation Strategies in CPathFMs}
\label{sec:methods}
\input{sections/methods.tex}

\vspace{-2mm}
\section{Evaluation Tasks}
\label{sec:tasks}

\input{sections/tasks.tex}

\vspace{-2mm}
\section{Future Directions}
\label{sec:future_work}
\input{sections/future_work.tex}

\section{Conclusion}
\label{sec:conclusion}
\input{sections/conclusion.tex}

\newpage
\appendix



\clearpage


\bibliographystyle{named}
\bibliography{ijcai25}

\end{document}

%% file: sections/introduction.tex
Histopathology with hematoxylin and eosin (H\&E) staining plays a fundamental role in disease diagnosis, prognosis, and treatment planning, particularly in oncology, where microscopic examination of tissue samples is critical for detecting pathological abnormalities. Traditionally, histopathological analysis relies on 
pathologists, who manually
examine
whole-slide images (WSIs) to identify disease patterns. However, this process is time-consuming, labor-intensive, and subject to inter-observer variability. With the increasing availability of digital pathology WSIs, deep learning-based computational pathology (CPath) models have emerged as a promising approach to enhancing diagnostic accuracy, reducing workload, and enabling large-scale pathology analysis. These models leverage convolutional neural networks (CNNs) and vision transformers (ViTs) to automate complex pathology tasks, such as tumor classification, biomarker discovery, and prognosis prediction.

In recent years, foundation models (FMs) have gained significant attention in CPath \cite{survey1}. Unlike other deep learning models that require large amounts of labeled data and are limited to specific tasks, computational pathology foundation models (CPathFMs) use larger models (such as various ViTs) as the backbone and are pre-trained on a vast and diverse set of unlabeled histopathological data through self-supervised learning (SSL). These models can be fine-tuned for various downstream pathology tasks using transfer learning, few-shot learning, or zero-shot learning, reducing the dependency on extensive expert manual annotations. 
Uni-modal CPathFMs are trained exclusively on histopathological images, capturing domain-specific features, while multi-modal CPathFMs, which integrate histopathological images with clinical linguistic data, enhance AI-driven pathology by leveraging complementary information available in electronic health records (EHRs). Despite some striking early successes, effectively pre-training CPathFMs remains challenging due to data limitations, adaptation difficulties, and evaluation inconsistencies.

\begin{figure*}[ht]
    \centering
    \includegraphics[width=0.9\linewidth]{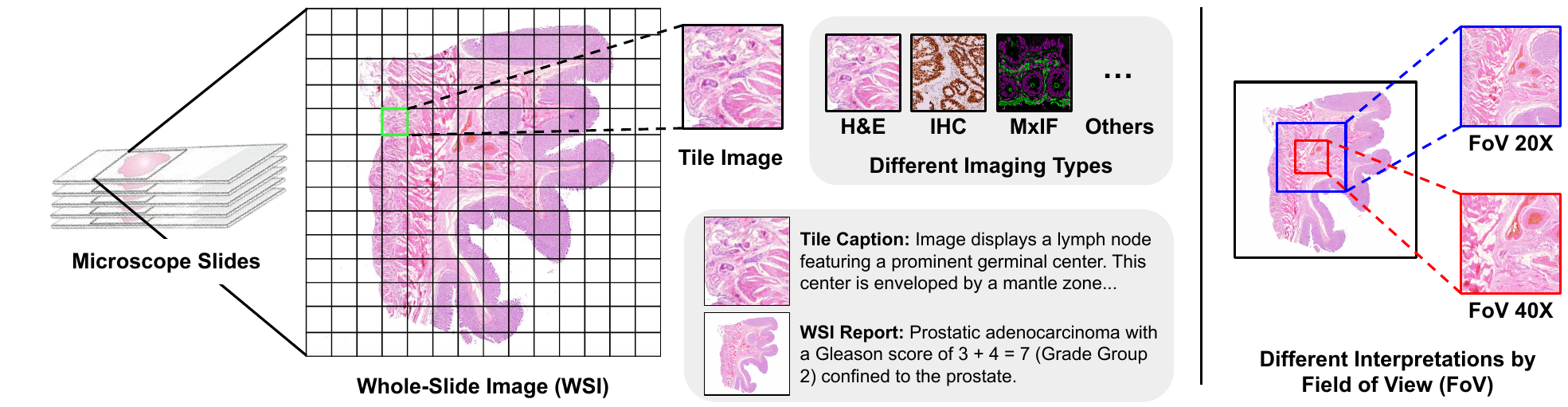}
    \vspace{-3mm}
    \caption{An illustrative example of data modalities and challenges in CPath. The figure illustrates different histopathology data types, including WSIs, tile images at multiple magnifications (Field of View, FoV), and imaging types (\textcolor{black}{H\&E, IHC, MxIF}). These elements are critical for developing CPathFMs, highlighting the complexity of multi-scale image representation and domain-specific challenges.}
    \label{fig:cpath}
    \vspace{-5.5mm}
\end{figure*}

The development of CPathFMs presents several key challenges. (1) Data challenges arise due to the high inter-institutional variability in histopathological images, including tissue sample preparation, differences in staining techniques,
and scanning resolutions. Labeled pathology datasets are often scarce and expensive to annotate, requiring expert pathologists for precise region-based annotations. Additionally, multi-modal data integration (\textit{e.g.,} combining WSIs with clinical text) remains complex due to challenges in alignment and feature fusion. (2) Adaptation challenges stem from the fact that, unlike natural image foundation models (\textit{e.g.,} CLIP \cite{CLIP} and DINO \cite{DINO}), CPathFMs require domain-specific fine-tuning to ensure robust generalization across different datasets. (3) Evaluation challenges further complicate CPathFM development, as the lack of standardized benchmarks and the diversity of evaluation tasks (\textit{e.g.,} classification, retrieval, generation, segmentation) make it difficult to assess model performance consistently across datasets and tasks.

While several survey papers have focused on CPathFMs, some of them emphasize benchmarking rather than a comprehensive investigation of existing methods \cite{benchmark1,benchmark2,benchmark3}. They cover too few approaches and lack a summary of pre-training datasets and evaluation tasks. Although \textit{Neidlinger} \textit{et al.,}~\shortcite{benchmark3} included a wide range of CPathFMs and pre-training datasets, their summarization of methods and datasets remains insufficiently detailed.
\textit{Ochi} \textit{et al.,}~\shortcite{survey1} and \textit{Chanda} \textit{et al.,}~\shortcite{survey2} conducted surveys on a large number of CPathFMs, but the methods they mentioned are not comprehensive and not up-to-date, and they did not provide a detailed introduction on how these methods are adapted to the pathology domain or distinguish between the different adaptations. Regarding summaries of evaluation tasks, the latter only listed the tasks without offering a systematic taxonomy, whereas the former's summary of the tasks was not comprehensive enough.


In this survey, we address these gaps by providing a comprehensive review of CPathFMs, with a focus on datasets, adaptation strategies, and evaluation tasks. Our main contributions include:
\begin{itemize}[leftmargin=*,itemsep=0pt, topsep=0pt]
    \item Providing an in-depth analysis of existing pathology datasets and data curation used for pre-training CPathFMs, identifying key challenges in generalization.
    \item Systematically reviewing adaptation techniques in pre-training CPathFMs, covering 28 existing and up-to-date models across both uni-modal (image-based) and multi-modal (image-text) paradigms.
    \item For the first time, thoroughly summarizing evaluation tasks, categorizing them into six main perspectives for assessing pre-trained CPathFMs.
    \item Identifying key future research directions, offering insights into the challenges and opportunities for advancing CPathFM development.
\end{itemize}

%% file: sections/background.tex




\subsection{Computational Pathology (CPath)}
\vspace{-1mm}
CPath is an interdisciplinary field that combines artificial intelligence, machine learning, and computer vision with digital pathology to enhance 
diagnosis, prognosis, and treatment planning. By leveraging whole-slide imaging and deep learning, CPath facilitates scalable and automated analysis of histopathological data, reducing reliance on manual examination by pathologists and improving diagnostic consistency. However, despite dramatic progress in CPathFMs and CPath more broadly, the deployment of machine learning methods in real-world clinical settings remains challenging, with many limitations in performance, usability, and regulatory compliance yet to be overcome.

Whole-slide imaging refers to the digital scanning of entire histopathology slides at high resolution, producing gigapixel WSIs that capture intricate tissue structures and morphologies. Analysis of WSIs is standard practice during manual image review, and is required by the FDA (U.S. Food and Drug Administration) for digital pathology applications, Due to their large size\footnote{The size of WSIs varies depending on factors such as tissue type and magnification level, scanning resolution, and file format, but they are typically gigapixel-scale images and can range from hundreds of megabytes to several gigabytes per slide.}, computational analysis WSIs  typically involves dividing them into smaller tile images. WSI reports summarize patient-level pathology interpretations and patient level data (\textit{e.g.,} tumor type and grade) but individuals tiles can also be annotated with tile captions, providing textual descriptions of specific tissue regions.

Figure \ref{fig:cpath} presents an illustration of key histopathology data modalities, including WSIs and tile images at multiple magnifications (Field of View, FoV), and different imaging techniques, such as H\&E, immunohistochemistry (IHC), and multiplex immunofluorescence (MxIF).
H\&E and IHC images are routinely used in clinical histopathology services and diagnosis but MxIF is a research technique still in development for diagnostic and clinical applications. All three types of images have been used in CPathFMs. The FoV plays a critical role in image interpretation, with tile-level analysis focusing on localized cellular structures, while WSI-level analysis provides a broader view of tissue morphology. Multi-modal CPathFMs further integrate histopathological images with clinical reports and captions, improving model generalizability and interpretability in AI-assisted pathology.

\begin{figure*}[t]
    \centering
    \includegraphics[width=0.90\linewidth]{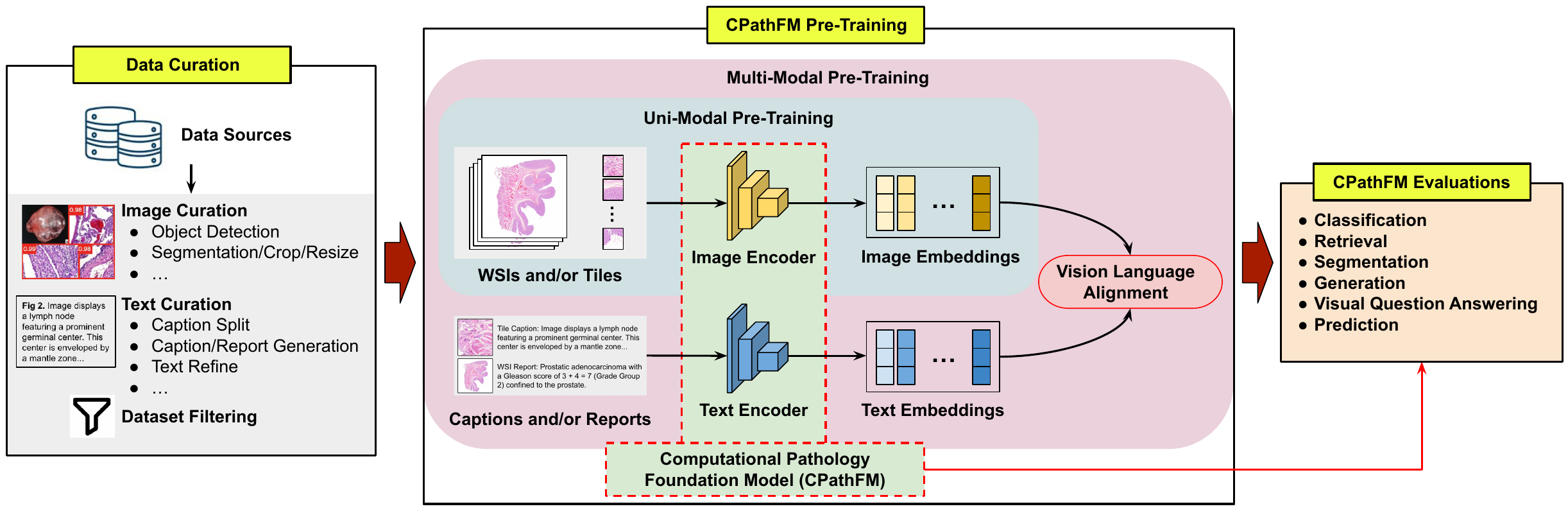}
    \vspace{-3mm}
    \caption{Overview of the pre-training pipeline for CPathFMs. The process involves data curation, including image curation, text curation, and dataset filtering, followed by uni-modal and multi-modal pre-training. The final CPathFMs are evaluated across multiple downstream tasks categorized into six main perspectives. }
    \label{fig:CPathFMs}
    \vspace{-6mm}
\end{figure*}

\subsection{ViT-Based SSL Frameworks for FMs}
\vspace{-1mm}

The ViT has emerged as a cornerstone for foundation models due to its powerful self-attention mechanism, which excels in capturing global dependencies. Its patch-based tokenization is inherently compatible with masked learning strategies, making it a natural fit for SSL. Furthermore, ViT’s scalability aligns with the large-scale nature of foundation models, and its Transformer architecture seamlessly integrates multiple modalities, such as vision and language. These properties have led to the development of various SSL frameworks that effectively leverage ViT, laying the groundwork for foundation models in pathology and beyond.

A key approach in SSL for ViTs is Masked Image Modeling (MIM), which is introduced by Masked Autoencoders (MAE)~\cite{MAE}. MIM leverages the patch-based structure of ViT to predict missing patches, thereby effectively learning rich and robust representations. Inspired by the success of BERT in NLP, BEiT~\cite{BEiT} incorporates a discrete tokenizer in MIM, while BEiT v2~\cite{Beitv2} refines the tokenization process through knowledge distillation based on BEiT.

In parallel, contrastive learning has played a crucial role in ViT-based SSL. DINO~\cite{DINO}, which follows a student-teacher paradigm, introduces a self-distillation approach that leverages ViT's global attention to learn robust features without explicit labels. DINOv2~\cite{DINOv2} enhances DINO by incorporating iBOT~\cite{iBOT}, which combines MIM with self-distillation. This integration allows DINOv2 to effectively capture fine-grained structures, improving its generalization across diverse datasets, and positioning it as a strong candidate for FMs.

Beyond vision-only learning, multi-modal contrastive frameworks such as CLIP~\cite{CLIP} integrate image and text encoders, enabling zero-shot classification and cross-modal retrieval. 
CoCa~\cite{CoCa} extends CLIP by incorporating a multi-modal decoder that generates language signals from visual signals, combining contrastive learning with image captioning and report synthesis. 
In terms of MIM, BEiT-3~\cite{BEiT-3} builds upon BEiT v2 by introducing more advanced token prediction mechanisms and leveraging multi-modal learning to extend its capabilities

\begin{table*}[ht]
\centering
\scriptsize
\setlength\tabcolsep{0.5pt}
\caption{Statistics of pathology datasets used for pre-training CPathFMs. The ``-'' symbol represents the absence of relevant information.}
\vspace{-3mm}
\begin{threeparttable}
\begin{tabular}{cccccccccccccccc}
\toprule
 & \multirow{2}{*}{\textbf{Reference}} & \multicolumn{2}{c}{\textbf{Data Description}\tnote{$\dagger$}} & \multirow{2}{*}{\textbf{Input Image Size}} & \textbf{Field of View} & \multicolumn{3}{c}{\textbf{Staining Types}} & \multicolumn{2}{c}{\textbf{Data Sources}} &\textbf{Corresponding}\\ 
\specialrule{0em}{0pt}{-0.2pt}
\cmidrule(lr){3-4} \cmidrule(lr){7-9} \cmidrule(lr){10-11}

\specialrule{0em}{-1pt}{0pt}
& & \textbf{\;\;\;\;\;\# WSIs\;\;\;\;\;} & \textbf{\# Tiles} &  & \textbf{(FoV)} & \textbf{H\&E} & \textbf{IHC} & \textbf{Others} & \textbf{Public} & \textbf{Private} & \textbf{Method}\\ 
\specialrule{0em}{-0.2pt}{0pt}
\midrule

\multirow{28}{*}{\makecell[c]{\rotatebox{90}{\textbf{Uni-modal}}}}  &\cite{CTransPath} & 32.2K & 15.6M & 1024$\times$1024 & 20$\times$ & \textcolor{green}{\ding{51}} & \textcolor{red}{\ding{55}} & \textcolor{red}{\ding{55}} & TCGA, PAIP & - & CTransPath\\ 
\specialrule{0em}{0pt}{-0.3pt}\cmidrule(lr){2-12}\specialrule{0em}{0pt}{-0.3pt}
 & \cite{REMEDIS} & 29.0K & 50.0M & 224$\times$224 & 20$\times$ & \textcolor{green}{\ding{51}} & \textcolor{red}{\ding{55}} & \textcolor{red}{\ding{55}} & TCGA & JFT54 &REMEDIS\\ \specialrule{0em}{0pt}{-0.3pt}\cmidrule(lr){2-12}\specialrule{0em}{0pt}{-0.3pt}
 & \cite{Lunit} & 36.7K & 32.6M & 512$\times$512 & \{20, 40\}$\times$ & \textcolor{green}{\ding{51}} & \textcolor{red}{\ding{55}} & \textcolor{red}{\ding{55}} & TCGA & TULIP &Lunit DINO\\ \specialrule{0em}{0pt}{-0.3pt}\cmidrule(lr){2-12}\specialrule{0em}{0pt}{-0.3pt}
 & \cite{Phikon} &6.1K & 43.4M & 224$\times$224 & 20$\times$ & \textcolor{green}{\ding{51}} & \textcolor{red}{\ding{55}} & \textcolor{red}{\ding{55}} & TCGA & - &Phikon\\ \specialrule{0em}{0pt}{-0.3pt}\cmidrule(lr){2-12}\specialrule{0em}{0pt}{-0.3pt}

 & \cite{Virchow}  & 1.5M & 2.0B & 224$\times$224 & 20$\times$ & \textcolor{green}{\ding{51}} & \textcolor{red}{\ding{55}} & \textcolor{red}{\ding{55}} & - & MSKCC  &Virchow\\ \specialrule{0em}{0pt}{-0.3pt}\cmidrule(lr){2-12}\specialrule{0em}{0pt}{-0.3pt}


 & \cite{campanella2024computational}  & 423K & 1.6B, 3.2B & 224$\times$224 & 20$\times$ & \textcolor{green}{\ding{51}} & \textcolor{red}{\ding{55}} & \textcolor{red}{\ding{55}} & - & MSHS &-\\ \specialrule{0em}{0pt}{-0.3pt}\cmidrule(lr){2-12}\specialrule{0em}{0pt}{-0.3pt}
 & \cite{RudolfV} & 134K & 1.2B &  256$\times$256 & \{20, 40, 80\}$\times$ & \textcolor{green}{\ding{51}} & \textcolor{green}{\ding{51}} & \textcolor{green}{\ding{51}} & TCGA & Proprietary  &RudolfV \\ \specialrule{0em}{0pt}{-0.3pt}\cmidrule(lr){2-12}\specialrule{0em}{0pt}{-0.3pt}
 & \cite{Kaiko} & 29.0K & 256M & 256$\times$256 & \{5, 10, 20, 40\}$\times$ & \textcolor{green}{\ding{51}} & \textcolor{red}{\ding{55}} & \textcolor{red}{\ding{55}} & TCGA & -  &Kaiko\\ 
 \specialrule{0em}{0pt}{-0.3pt}\cmidrule(lr){2-12}\specialrule{0em}{0pt}{-0.3pt}
 & \cite{PLUTO} & 158.8K & 195M &  224$\times$224 & \{20, 40\}$\times$ & \textcolor{green}{\ding{51}} & \textcolor{green}{\ding{51}} & \textcolor{green}{\ding{51}} & TCGA, \textit{etc.} &Proprietary &PLUTO\\
 \specialrule{0em}{0pt}{-0.3pt}\cmidrule(lr){2-12}\specialrule{0em}{0pt}{-0.3pt}
 & \cite{GigaPath} & 171K & 1.4B &  256$\times$256 & 20$\times$ & \textcolor{green}{\ding{51}} & \textcolor{green}{\ding{51}} & \textcolor{red}{\ding{55}} &  - &  PHS&GigaPath\\ \specialrule{0em}{0pt}{-0.3pt}\cmidrule(lr){2-12}\specialrule{0em}{0pt}{-0.3pt}
 & \cite{Hibou} & 1.1M & 512M, 1.2B &  224$\times$224 & 20$\times$ & \textcolor{green}{\ding{51}} & \textcolor{green}{\ding{51}} & \textcolor{green}{\ding{51}} &  - & Proprietary&Hibou\\ 
 \specialrule{0em}{0pt}{-0.3pt}\cmidrule(lr){2-12}\specialrule{0em}{0pt}{-0.3pt}
 & \cite{BEPH} & 11.7K & 11.7M &  224$\times$224 & 20$\times$ & 
 \textcolor{green}{\ding{51}} & \textcolor{red}{\ding{55}}  & \textcolor{red}{\ding{55}}  & TCGA & - &BEPH\\\specialrule{0em}{0pt}{-0.3pt}\cmidrule(lr){2-12}\specialrule{0em}{0pt}{-0.3pt}
 & \cite{GPFM} & 72.3K &  190.2M & 512$\times$512 & - & 
 \textcolor{green}{\ding{51}} & \textcolor{red}{\ding{55}} & \textcolor{red}{\ding{55}} & TCGA, GTEx, \textit{etc.}  & - & GPFM\\\specialrule{0em}{0pt}{-0.3pt}\cmidrule(lr){2-12}\specialrule{0em}{0pt}{-0.3pt}
 & \cite{Virchow2} & 3.1M & 2.0B &  392$\times$392 & \{5, 10, 20, 40\}$\times$ & \textcolor{green}{\ding{51}}  & \textcolor{green}{\ding{51}} & \textcolor{red}{\ding{55}} & - & MSKCC&Virchow2\\ \specialrule{0em}{0pt}{-0.3pt}\cmidrule(lr){2-12}\specialrule{0em}{0pt}{-0.3pt}
 & \cite{Phikon-v2} & 58.0K & 456M & 224$\times$224 & 20$\times$ & \textcolor{green}{\ding{51}} & \textcolor{green}{\ding{51}} & \textcolor{green}{\ding{51}} & TCGA, GTEx, \textit{etc.} & Proprietary$\times$4 &Phikon-v2\\
 \specialrule{0em}{0pt}{-0.3pt}\cmidrule(lr){2-12}\specialrule{0em}{0pt}{-0.3pt}

 & \cite{UNI} & 100K & 100M & 256$\times$256, 512$\times$512 & 20$\times$ & \textcolor{green}{\ding{51}} & \textcolor{red}{\ding{55}} & \textcolor{red}{\ding{55}} &  GTEx & MGH, BWH &UNI\\ 
 \specialrule{0em}{0pt}{-0.3pt}\cmidrule(lr){2-12}\specialrule{0em}{0pt}{-0.3pt}
 
 & \cite{H-optimus-0} & 500K+ & 100M+ & - & - & \textcolor{green}{\ding{51}} & \textcolor{red}{\ding{55}} & \textcolor{red}{\ding{55}} &  - & Proprietary &H-optimus-0\\ 
 \specialrule{0em}{0pt}{-0.3pt}\cmidrule(lr){2-12}\specialrule{0em}{0pt}{-0.3pt}
 & \cite{Atlas} & 1.2M & 3.4B &  256$\times$256 & \{5, 10, 20, 40\}$\times$ & \textcolor{green}{\ding{51}} & \textcolor{green}{\ding{51}} & \textcolor{green}{\ding{51}} & - & Proprietary  &Atlas \\

\midrule
\midrule

\multirow{23}{*}{\makecell[c]{\rotatebox{90}{\textbf{Multi-modal}}}}  & \cite{PLIP} & \multicolumn{2}{c}{208K Tile-Caption Pairs} & 224$\times$224 & - & \textcolor{green}{\ding{51}} & \textcolor{green}{\ding{51}} & - & Twitter, PathLAION & - & PLIP\\
\specialrule{0em}{0pt}{-0.3pt}\cmidrule(lr){2-12}\specialrule{0em}{0pt}{-0.3pt}
 & \cite{PathCLIP} & \multicolumn{2}{c}{207K Tile-Caption Pairs} & - & - &\textcolor{green}{\ding{51}} & \textcolor{green}{\ding{51}} & \textcolor{red}{\ding{55}} & PMC OA & LBC&PathCLIP\\ 
\specialrule{0em}{0pt}{-0.3pt}\cmidrule(lr){2-12}\specialrule{0em}{0pt}{-0.3pt}
 & \cite{QuiltNet} & \multicolumn{2}{c}{438K Tiles and 802K Captions} & Avg. 882$\times$1648 & \{10-40\}$\times$ & \textcolor{green}{\ding{51}} & \textcolor{green}{\ding{51}} & - & YouTube, Twitter, \textit{etc.} & - & QuiltNet\\

\specialrule{0em}{0pt}{-0.3pt}\cmidrule(lr){2-12}\specialrule{0em}{0pt}{-0.3pt}
 & \multirow{2}{*}{\cite{CONCH}} & \multicolumn{2}{c}{21K WSIs from 16M Tile Images} & 256$\times$256  & 20$\times$ & \textcolor{green}{\ding{51}} &  - &- & - &Proprietary  &\multirow{2}{*}{CONCH}\\
& & \multicolumn{2}{c}{1.17M Tile-Caption Pairs} & 448$\times$448 & - & \textcolor{green}{\ding{51}} & \textcolor{green}{\ding{51}} & \textcolor{green}{\ding{51}} & PMC OA & EDU& \\ 
\specialrule{0em}{0pt}{-0.3pt}\cmidrule(lr){2-12}\specialrule{0em}{0pt}{-0.3pt}
 & \cite{PRISM} & \multicolumn{2}{c}{587K WSIs and 195K Reports} & 224$\times$224 & 20$\times$  & \textcolor{green}{\ding{51}} & \textcolor{red}{\ding{55}} & \textcolor{red}{\ding{55}} & TCGA & Proprietary&PRISM \\ 
\specialrule{0em}{0pt}{-0.3pt}\cmidrule(lr){2-12}\specialrule{0em}{0pt}{-0.3pt}

  & \cite{CHIEF} & \multicolumn{2}{c}{60K WSI-Label Pairs} & 256$\times$256 & 10$\times$ & \textcolor{green}{\ding{51}} & \textcolor{red}{\ding{55}} & \textcolor{red}{\ding{55}} & TCGA, GTEx, \textit{etc.} & Proprietary &CHIEF\\ 
\specialrule{0em}{0pt}{-0.3pt}\cmidrule(lr){2-12}\specialrule{0em}{0pt}{-0.3pt}

& \multirow{2}{*}{\cite{KEP}} & \multicolumn{2}{c}{A KG with 50.5K Pathology Attributes} & - & - & - & - & - & OncoTree, \textit{etc.} & - & \multirow{2}{*}{KEP}\\

&  & \multicolumn{2}{c}{576.6K, 138.9K Tile-Caption Pairs} & 224$\times$224 & - & \textcolor{green}{\ding{51}} & \textcolor{green}{\ding{51}} & - & Quilt-1M, OpenPath & - & \\

\specialrule{0em}{0pt}{-0.3pt}\cmidrule(lr){2-12}\specialrule{0em}{0pt}{-0.3pt}
 & \multirow{3}{*}{\cite{TITAN}}&  \multicolumn{2}{c}{336K WSIs} & 512$\times$512 & 20$\times$ & \textcolor{green}{\ding{51}}  &  \textcolor{green}{\ding{51}} & \textcolor{red}{\ding{55}} & GTEx &  Proprietary &\multirow{3}{*}{TITAN}\\ 
& & \multicolumn{2}{c}{423K Tile-Caption Pairs} & 8192$\times$8192 & 20$\times$ & \textcolor{green}{\ding{51}}  &  \textcolor{green}{\ding{51}} & \textcolor{red}{\ding{55}} & GTEx &  Proprietary & \\ 
& &\multicolumn{2}{c}{183K WSI-Report Pairs} &32768$\times$32768 & - &\textcolor{green}{\ding{51}} &\textcolor{green}{\ding{51}} & \textcolor{red}{\ding{55}} & GTEx &  Proprietary& \\ 
\specialrule{0em}{0pt}{-0.3pt}\cmidrule(lr){2-12}\specialrule{0em}{0pt}{-0.3pt}

& \multirow{2}{*}{\cite{KEEP}} & \multicolumn{2}{c}{A KG with 139K Disease Attributes} & - & - & - & - & - & DO, UMLS & - & \multirow{2}{*}{KEEP}\\

&  & \multicolumn{2}{c}{143K Tile-Caption Pairs} & 224$\times$224 & - & \textcolor{green}{\ding{51}} & \textcolor{green}{\ding{51}} & - & Quilt-1M, OpenPath & - & \\
\specialrule{0em}{0pt}{-0.3pt}\cmidrule(lr){2-12}\specialrule{0em}{0pt}{-0.3pt}

& \multirow{2}{*}{\cite{MUSK}} & \multicolumn{2}{c}{1B Text Tokens and 50M Tiles} & 384$\times$384 & \{10, 20, 40\}$\times$ & -\textcolor{green}{\ding{51}} & \textcolor{red}{\ding{55}} & \textcolor{red}{\ding{55}} & PMC OA, TCGA & - & \multirow{2}{*}{MUSK}\\

&  & \multicolumn{2}{c}{1M Tile-Caption Pairs} & 384$\times$384 & 20$\times$ & \textcolor{green}{\ding{51}} & \textcolor{red}{\ding{55}} & \textcolor{red}{\ding{55}} & Quilt-1M, PathCap & - & \\

\bottomrule
\end{tabular}
\begin{tablenotes}
\item[$\dagger$] The pre-training data for uni-modal CPathFMs primarily consists of the number of WSIs and tiles. However, the situation is more complex for multi-modal models, so we provide a textual description for clarification.
    \end{tablenotes}
\end{threeparttable}

\label{tab:datasets}
\vspace{-6mm}
\end{table*}

\subsection{Challenges in Pre-training CPathFMs}
\vspace{-1mm}
While self-supervised contrastive learning frameworks have significantly advanced the development of CPathFMs, pre-training these models remains challenging due to issues in data availability, adaptation strategies, and evaluation complexities that are closely tied to the entire pre-training process.
As shown in Figure \ref{fig:CPathFMs}, pre-training CPathFMs involves three key steps: first, preparing the pre-training dataset through data curation; second, training within an adapted SSL framework;
and finally, evaluating the model on a series of downstream tasks.
Addressing limitations during pre-training is crucial to improving the generalizability and clinical applicability of CPathFMs.

A primary challenge in pre-training CPathFMs is data availability. Histopathology datasets are often restricted by ethical approvals and access permissions, limiting the availability of large, diverse datasets. Public datasets are rare, often from a single institution, reducing diversity. Additionally, WSIs are large, gigapixel-scale images that pose storage and computational challenges, requiring advanced data management solutions. Annotation is another major obstacle, as pathology image labeling requires expert knowledge, often involving multiple pathologists to ensure accuracy. This process is time-consuming and costly, leading to a shortage of well-annotated datasets. Furthermore, pathology data exhibit substantial variability in staining protocols, magnification levels, and organ-specific structures, introducing domain shifts that complicate model generalization. 
Data imbalance further exacerbates these issues, as rare disease types account for only a small fraction of available data, causing models to overfit common diseases while performing poorly on underrepresented conditions.

Beyond data-related issues, adapting pre-trained models to diverse pathology tasks presents additional challenges. Unlike natural image datasets, which are often well-curated and standardized, pathology images exhibit significant inter-institutional variability, requiring pre-training strategies that can effectively capture and generalize across this diversity. This demand for higher generalization ability makes it difficult to develop CPathFMs that perform robustly across different patient populations and imaging conditions. Another key limitation is the difficulty of processing gigapixel-scale WSIs, which are too large for traditional deep learning architectures to handle directly. Many current approaches rely on tile-based learning, where WSIs are broken into smaller patches, but this often leads to context fragmentation, making it difficult for models to retain spatial relationships within the tissue. Efficient multi-scale learning techniques are needed to bridge this gap.

Evaluation presents a significant challenge in pre-training CPathFMs. The wide range of downstream pathology tasks, including classification, retrieval, segmentation, generation, and so on, complicates performance assessment, making it difficult to compare CPathFMs across different institutions and tasks. The heterogeneity in evaluation methodologies further hinders the establishment of a universal benchmarking framework, limiting the ability to systematically assess and compare model performance.


%% file: sections/pretraining_datasets.tex
Although early CPathFMs used relatively small and homogeneous pre-training datasets, recent studies have shown that higher quality, larger scale, and more diverse pathology pre-training datasets are more beneficial for adapting the foundation models trained on natural image datasets or existing SSL frameworks to the pathology domain ~\cite{Virchow2}. 
Therefore, summarizing the datasets used for pre-training CPathFMs can provide valuable insight into requirements for future research on CPathFMs. Table \ref{tab:datasets} provides a summary of pre-training datasets utilized by each method discussed in Section \ref{sec:methods}.









 
Most CPathFMs utilize multiple data sources to construct larger and more diverse pathology pre-training datasets. Aside from inaccessible private data, existing methods often acquire WSIs from large-scale public pathology datasets such as The Cancer Genome Atlas (TCGA)
\cite{TCGA}, The Genotype-Tissue Expression (GTEx) Consortium
\cite{GTEx}, 
as well as tile-caption pairs from the PubMed Central Open Access Dataset (PMC OA)
\cite{CONCH}. A common method for multi-modal pathology data is to collect it from the internet, for example, \textit{Ikezogwo et al.,}~\shortcite{QuiltNet} curate a large-scale dataset named Quilt-1M from multiple different online sources. \textit{Huang et al.,}~\cite{PLIP} and \textit{Sun et al.,}~\cite{PathCLIP} also released their pretraining datasets, OpenPath and PathCap, respectively, alongside the proposal of CPathFM.

The diversity of data sources imposes higher requirements on data curation. Although the processes vary across methods, some steps are similar. As shown in Figure \ref{fig:CPathFMs}, images are typically subject to detection and segmentation of subfigures; resizing and cropping to meet model requirements, or using off-the-shelf LLMs to generate captions or reports for multimodal task needs. For text, titles are usually segmented or refined using LLMs. The cross-modal curation process includes the alignment of subfigures and subcaptions, as well as the construction of instructions. Finally, the data is filtered to select the required pathology data.



From the perspective of data types, the pre-training datasets for uni-modal CPathFMs are relatively straightforward, typically containing WSIs and the tiles extracted from them. 
However, multi-modal CPathFMs often use different data types as inputs depending on the training needs. Tile-caption pairs are typically used for tile-level, while WSI-report pairs are generally employed for slide-level (note that one or more WSIs usually have a corresponding clinical report in practice). 
Notably, \textit{Wang} \textit{et al.,}~\shortcite{CHIEF} used the WSI labels (anatomical sites) as the textual information to construct the data pairs. 

%% file: sections/methods.tex
\begin{table*}[ht]
\centering
\scriptsize
\setlength{\abovecaptionskip}{0pt}
\setlength{\belowcaptionskip}{2pt}
\setlength\tabcolsep{0pt}
\begin{threeparttable}
\caption{Overview of architecture and adaptation strategies of CPathFMs}
\vspace{-8mm}
\label{tab:methods}
\begin{tabular}{ccccccccccc}
\toprule
\multirow{2}{*}{\textbf{}} & \multirow{2}{*}{\textbf{Model}} & \multirow{2}{*}{\textbf{Reference}} &\textbf{SSL}  & \multicolumn{2}{c}{\textbf{Backbone [\# Param.]\tnote{$\dagger$}}} & \textbf{Input } & \multicolumn{2}{c}{\textbf{Pre-training Strategy\tnote{$\ddagger$}}} & \textbf{Model}\\ 
\specialrule{0em}{0pt}{-0.2pt}
\cmidrule(lr){5-6} \cmidrule(lr){8-9}
\specialrule{0em}{-1pt}{0pt}

 &  &  & \textbf{Framework} & \textbf{Vision} & \textbf{Language}  & \textbf{Images} &  \textbf{\;\;Vision} & \textbf{Language} & \textbf{Availability}\\ 

\midrule
\multirow{30}{*}{\makecell[c]{\rotatebox{90}{\textbf{Uni-modal}}}} & CTransPath & \cite{CTransPath} & MoCo v3* & Swin Transformer [\textcolor{orange}{28M}] & - & Tiles & S & - & \textcolor{green}{\ding{51}}\\ \specialrule{0em}{0pt}{-0.3pt}\cmidrule(lr){2-10}\specialrule{0em}{0pt}{-0.3pt}
    & REMEDIS & \cite{REMEDIS} & SimCLR  &  ResNet-152 (2$\times$) [\textcolor{orange}{232M}] & - & Tiles & D & -& \textcolor{green}{\ding{51}}\\ \specialrule{0em}{0pt}{-0.3pt}\cmidrule(lr){2-10}\specialrule{0em}{0pt}{-0.3pt}
    & Lunit DINO & \cite{Lunit} & DINO  & ViT-S/(8,16) [\textcolor{orange}{22M}] & - & Tiles & S & - & \textcolor{green}{\ding{51}}\\ \specialrule{0em}{0pt}{-0.3pt}\cmidrule(lr){2-10}\specialrule{0em}{0pt}{-0.3pt}
    & Phikon & \cite{Phikon} & iBOT & ViT-B/16 [\textcolor{orange}{86M}] & - & Tiles & S & - & \textcolor{red}{\ding{55}}\\
    \specialrule{0em}{0pt}{-0.3pt}\cmidrule(lr){2-10}\specialrule{0em}{0pt}{-0.3pt}

    & Virchow & \cite{Virchow} & DINOv2 & ViT-H/14 [\textcolor{orange}{632M}] & - & Tiles  & S & -& \textcolor{green}{\ding{51}}\\ \specialrule{0em}{0pt}{-0.3pt}\cmidrule(lr){2-10}\specialrule{0em}{0pt}{-0.3pt}
    & \multirow{2}{*}{-} & \multirow{2}{*}{\cite{campanella2024computational}} & DINO & ViT-S [\textcolor{orange}{22M}], ViT-B [\textcolor{orange}{86M}] & - & \multirow{2}{*}{Tiles}  & \multirow{2}{*}{S} & - & \multirow{2}{*}{\textcolor{red}{\ding{55}}}\\
    & & & MAE & ViT-L [\textcolor{orange}{307M}] & - & && -\\ \specialrule{0em}{0pt}{-0.3pt}\cmidrule(lr){2-10}\specialrule{0em}{0pt}{-0.3pt}
    & RudolfV & \cite{RudolfV} & DINOv2 & ViT-L/14 [\textcolor{orange}{307M}] & - & Tiles  & D & - & \textcolor{red}{\ding{55}}\\ \specialrule{0em}{0pt}{-0.3pt}\cmidrule(lr){2-10}\specialrule{0em}{0pt}{-0.3pt}
    & \multirow{2}{*}{Kaiko} & \multirow{2}{*}{\cite{Kaiko}} & DINO & ViT-S/(8,16) [\textcolor{orange}{22M}], ViT-B/(8,16) [\textcolor{orange}{86M}] & - & \multirow{2}{*}{Tiles}  & \multirow{2}{*}{D}& \multirow{2}{*}{-} & \multirow{2}{*}{\textcolor{green}{\ding{51}}}\\
    &  & & DINOv2 & ViT-L/14 [\textcolor{orange}{307M}] & -&   &\\ \specialrule{0em}{0pt}{-0.3pt}\cmidrule(lr){2-10}\specialrule{0em}{0pt}{-0.3pt}
    & PLUTO & \cite{PLUTO} &  DINOv2* & FlexiViT-S [\textcolor{orange}{22M}] & -& Tiles & S & -& \textcolor{green}{\ding{51}}\\ \specialrule{0em}{0pt}{-0.3pt}\cmidrule(lr){2-10}\specialrule{0em}{0pt}{-0.3pt}
    & GigaPath & \cite{GigaPath} & DINOv2* & ViT-G/14 [\textcolor{orange}{1.1B}] \& LongNet [\textcolor{orange}{125M}] & -& WSIs & S, S & -& \textcolor{green}{\ding{51}}\\ \specialrule{0em}{0pt}{-0.3pt}\cmidrule(lr){2-10}\specialrule{0em}{0pt}{-0.3pt}
    & Hibou & \cite{Hibou} & DINOv2 & ViT-B/14 [\textcolor{orange}{86M}], ViT-L/14 [\textcolor{orange}{307M}] & -& Tiles & S & - & \textcolor{green}{\ding{51}}\\ 
    \specialrule{0em}{0pt}{-0.3pt}\cmidrule(lr){2-10}\specialrule{0em}{0pt}{-0.3pt}
    & BEPH & \cite{BEPH} & BEiTv2 & ViT-B/16 [\textcolor{orange}{86M}] \& VQ-KD [\textcolor{orange}{86M}] & - & Tiles & D, D & - & \textcolor{green}{\ding{51}}\\ \specialrule{0em}{0pt}{-0.3pt}\cmidrule(lr){2-10}\specialrule{0em}{0pt}{-0.3pt}
    & GPFM & \cite{GPFM} & DINOv2* & ViT-L [\textcolor{orange}{307M}] & - & Tiles & S & - & \textcolor{green}{\ding{51}}\\ \specialrule{0em}{0pt}{-0.3pt}\cmidrule(lr){2-10}\specialrule{0em}{0pt}{-0.3pt}
    & Virchow2 & \cite{Virchow2} & DINOv2* & ViT-B/16 [\textcolor{orange}{86M}] & - & Tiles & S & - & \textcolor{green}{\ding{51}}\\  \specialrule{0em}{0pt}{-0.3pt}\cmidrule(lr){2-10}\specialrule{0em}{0pt}{-0.3pt}
    & Phikon-v2 & \cite{Phikon-v2} & DINOv2 & ViT-L/16 [\textcolor{orange}{307M}] & - & Tiles & S& - & \textcolor{red}{\ding{55}}\\ \specialrule{0em}{0pt}{-0.3pt}\cmidrule(lr){2-10}\specialrule{0em}{0pt}{-0.3pt}

    & UNI & \cite{UNI} & DINOv2 & ViT-L/16 [\textcolor{orange}{307M}] & - & Tiles  & S & -& \textcolor{green}{\ding{51}}\\ 
    \specialrule{0em}{0pt}{-0.3pt}\cmidrule(lr){2-10}\specialrule{0em}{0pt}{-0.3pt}

    & H-optimus-0 & \cite{H-optimus-0} & DINOv2 & ViT-G/14 [\textcolor{orange}{1.1B}] & - & Tiles  & S & -& \textcolor{green}{\ding{51}}\\ 
    \specialrule{0em}{0pt}{-0.3pt}\cmidrule(lr){2-10}\specialrule{0em}{0pt}{-0.3pt}

    & Atlas & \cite{Atlas} & DINOv2 & ViT-H/14 [\textcolor{orange}{632M}] & - & Tiles  & D & -& \textcolor{red}{\ding{55}}\\ 
    
    \midrule
    \midrule

\multirow{18}{*}{\makecell[c]{\rotatebox{90}{\textbf{Multi-modal}}}} & PLIP & \cite{PLIP} & CLIP & ViT-B/32 [\textcolor{orange}{86M}] & Transformer Layers [\textcolor{orange}{63M}] & Tiles & D & D & \textcolor{green}{\ding{51}}\\ \specialrule{0em}{0pt}{-0.3pt}\cmidrule(lr){2-10}\specialrule{0em}{0pt}{-0.3pt}
    & PathCLIP & \cite{PathCLIP}& CLIP &  ViT-B/32 [\textcolor{orange}{86M}] & Transformer Layers [\textcolor{orange}{63M}] & Tiles & D & D & \textcolor{red}{\ding{55}}\\ 
    
    \specialrule{0em}{0pt}{0.2pt}
    
    \specialrule{0em}{0pt}{-0.3pt}\cmidrule(lr){2-10}\specialrule{0em}{0pt}{-0.3pt}

    & QuiltNet & \cite{QuiltNet} & CLIP & ViT-B/(16,32) [\textcolor{orange}{86M}]&GPT-2 [\textcolor{orange}{1.5B}] \& PubMedBERT [\textcolor{orange}{100M}] & Tiles & D & D & \textcolor{green}{\ding{51}}\\ \specialrule{0em}{0pt}{-0.3pt}\cmidrule(lr){2-10}\specialrule{0em}{0pt}{-0.3pt}
    
    & CONCH & \cite{CONCH} & CoCa, iBOT & ViT-B/16 [\textcolor{orange}{86M}] & Transformer Layers [\textcolor{orange}{$\sim$86M}] & Tiles & S & D & \textcolor{green}{\ding{51}}\\
    \specialrule{0em}{0pt}{-0.3pt}\cmidrule(lr){2-10}\specialrule{0em}{0pt}{-0.3pt}
    & PRISM & \cite{PRISM} & CoCa & ViT-H/14 [\textcolor{blue!80}{632M}] \& Perceiver Net. [\textcolor{orange}{105M}] & BioGPT [\textcolor{orange}{345M}, \textcolor{blue!80}{172M}] & WSIs & F, S & D, F& \textcolor{green}{\ding{51}}\\ \specialrule{0em}{0pt}{-0.3pt}\cmidrule(lr){2-10}\specialrule{0em}{0pt}{-0.3pt} 

    
    & CHIEF & \cite{CHIEF} & CLIP* & Swin Transformer [\textcolor{orange}{28M}] & Transformer Layers [\textcolor{orange}{63M}] & WSIs & D & D & \textcolor{green}{\ding{51}}\\ \specialrule{0em}{0pt}{-0.3pt}\cmidrule(lr){2-10}\specialrule{0em}{0pt}{-0.3pt}

    & KEP & \cite{KEP} & CLIP* &  ViT-B/(16,32) [\textcolor{orange}{86M}] & PubMedBERT [\textcolor{orange}{100M}] & Tiles & D & S & \textcolor{green}{\ding{51}}\\
    \specialrule{0em}{0pt}{-0.3pt}\cmidrule(lr){2-10}\specialrule{0em}{0pt}{-0.3pt} 
    
    & TITAN & \cite{TITAN} & CoCa, iBOT* &  ViT-B/16 [\textcolor{blue!80}{86M}] \& ViT-S [\textcolor{orange}{22M}] & Transformer Layers [\textcolor{orange}{$\sim$86M}]  & WSIs & F, S & D & \textcolor{green}{\ding{51}}\\
    \specialrule{0em}{0pt}{-0.3pt}\cmidrule(lr){2-10}\specialrule{0em}{0pt}{-0.3pt} 

    & KEEP & \cite{KEEP} & CLIP* & ViT-L [\textcolor{orange}{307M}]  & PubMedBERT [\textcolor{orange}{100M}] & Tiles & D & S & \textcolor{green}{\ding{51}}\\
    \specialrule{0em}{0pt}{-0.3pt}\cmidrule(lr){2-10}\specialrule{0em}{0pt}{-0.3pt}

    & \multirow{2}{*}{MUSK} & \multirow{2}{*}{\cite{MUSK}} & \multirow{2}{*}{CoCa*, BEiT-3} &  V-FFN [\textcolor{orange}{202M}]  &  L-FFN [\textcolor{orange}{202M}] & \multirow{2}{*}{Tiles} & \multirow{2}{*}{S} & \multirow{2}{*}{S} & \multirow{2}{*}{\textcolor{green}{\ding{51}}}\\

    &  & &  &\multicolumn{2}{c}{Shared Attention Layers [\textcolor{orange}{202M}]} &  &  &  & \\

\bottomrule
\end{tabular}

\begin{tablenotes}
        \item[*] Made domain-specific improvements or extensions to the SSL framework for pathology.
        \item[$\dagger$] For simplicity, we have streamlined some expressions. For example, ``/8" denotes a patch size of 8$\times$8 pixels, and ``/(8,16)" represents ``/8" and ``/16", respectively. \\ The \textbf{\textcolor{orange}{orange}} color in [ ] represents the parameters that are being trained or tuned, while the \textbf{\textcolor{blue!80}{blue}} color represents the frozen parameters.
\item[$\ddagger$] Pre-training strategies: \textbf{F}: Frozen, \textbf{S}: From Scratch, \textbf{D}: Domain-Specific Tuning.
\end{tablenotes}
\end{threeparttable}
\vspace{-5.5mm}
\end{table*}
SSL has been widely applied in the development of CPathFMs to address the lack of labels. These models typically adapt SSL frameworks that have proven successful in natural images, and perform pre-training on carefully curated pathology datasets. Depending on the type of pathology data they used, these approaches can be categorized into uni-modal and multi-modal methods, as introduced in Table~\ref{tab:methods}.


\subsection{Uni-Modal CPathFMs}
\vspace{-1mm}

Uni-modal CPathFMs are generally trained on large, domain-specific pathology datasets using SSL frameworks to learn robust representations of pathological images without labeled data. Although there are some MIM-based methods, self-supervised contrastive learning methods play a dominant role.
Similar to the development of contrastive learning in natural images, CPathFMs were initially proposed within the MoCo~\cite{MoCo3} and SimCLR~\cite{SimCLR} frameworks. 
Following a transition through the DINO, DINOv2 was established as the leading framework, serving as the foundation for numerous subsequent studies.

\textbf{DINO-based CPathFMs.}
As a successful application of SSL on ViT, DINO has been adopted as a framework for training CPathFMs. 
%
\textit{Campanella}~\textit{et al.,}~\shortcite{campanella2024computational} compared the performance of DINO and MAE on different scales of pathology datasets, 
ultimately demonstrating the superiority of DINO for pre-training CPathFMs. 
\textit{Kang}~\textit{et al.,}~\shortcite{Lunit} focused on domain-aligned pre-training and proposed data augmentation and curation strategies specifically for pathological images.

\textbf{DINOv2-based CPathFMs.} Most studies using the DINOv2 framework, such as UNI \cite{UNI}, mainly focus on larger ViT models and more extensive and diverse pre-training datasets. Among them, RudolfV \cite{RudolfV} incorporates the domain knowledge of pathologists in dataset construction. 
Some methods have developed pathology-adapted training methods within the DINOv2 framework, such as Kaiko~\cite{Kaiko} which develops the Online Patching method during the pre-training process, allowing for the online high-throughput extraction of patches of arbitrary size and resolution. Virchow2 \cite{Virchow2} replaces the original entropy estimator in DINOv2 with the kernel density estimator (KDE) and PLUTO~\cite{PLUTO} enhances the DINOv2 loss by incorporating a MAE objective and a Fourier-based loss term to regulate the retention of low- and high-frequency components. Additionally, GPFM~\cite{GPFM} selects multiple existing CPathFMs as expert models and builds a unified knowledge distillation framework by performing Expert Knowledge Distillation on top of DINOv2.
Unlike previous methods, GigaPath~\cite{GigaPath} focuses on learning the representation of whole-slide images. It first uses the tile-level encoder, trained under the DINOv2 framework, to learn the representation of a sequence of tiles obtained by dividing the whole slide, treating them as visual tokens. These tokens are then fed into LongNet \cite{LongNet}, which utilizes its Dilated Attention mechanism to perform sparse attention computation, thereby obtaining the overall slide representation.

\textbf{Other Uni-Modal CPathFMs.}
While the majority of uni-modal methods focus on DINO and DINOv2, some methods employ other SSL frameworks. CTransPath \cite{CTransPath} adds a branch to MoCov3 to generate queries that retrieve semantically similar samples from the memory bank as positive samples, thus guiding the network's training with a semantically relevant contrastive loss.
REMEDIS \cite{REMEDIS} transfers a ResNet model, pre-trained on large-scale natural images, to the SimCLR framework for self-supervised training on pathological images.
Additionally, Phikon~\cite{Phikon} and BEPH~\cite{BEPH} directly train a ViT model within the MIM-based SSL framework iBOT and BEiTv2, respectively.

\subsection{Multi-Modal CPathFMs}
\vspace{-1mm}

Multi-modal CPathFMs enhance the model’s understanding of pathological images by aligning paired image-text data under the visual-language multi-modal SSL frameworks, such as CLIP and CoCa. 
These methods typically train pre-trained uni-modal modules using uni-modal SSL frameworks before performing joint visual-language pre-training, which has been shown to improve the performance of downstream tasks \cite{CONCH}.

\vspace{-1mm}
\textbf{CLIP-based CPathFMs.}
The success of CLIP on natural images has inspired research aimed at applying it in the pathology domain, where it enhances model interpretability for pathology AI by leveraging paired histopathological images and textual descriptions (\textit{e.g.,} pathology reports and expert annotations). PLIP \cite{PLIP}, PathCLIP \cite{PathCLIP} and QuiltNet~\cite{QuiltNet} all fine-tune a pre-trained CLIP model using datasets composed of paired tiles and their captions. 
CHIEF \cite{CHIEF} uses an image encoder from CTransPath to encode the tile sequence extracted from WSIs to obtain WSI-level features and encodes anatomical site information as textual features using the original CLIP's text encoder. The two are combined to obtain rich WSI-level multi-modal representations.
To integrate domain-specific knowledge, \textit{Zhou et al.,}~\shortcite{KEP} constructed a pathology knowledge graph (KG) and encode it using a knowledge encoder, which then guides visual-language pretraining.
Similarly, KEEP~\cite{KEEP} builds a disease KG for encoding and employs knowledge-guided dataset structuring to generate tile-caption pairs for pretraining within the CLIP framework, incorporating positive mining, hardest negative sampling, and false negative elimination strategies.


\begin{figure*}[t!]
    \centering
    \includegraphics[width=\linewidth]{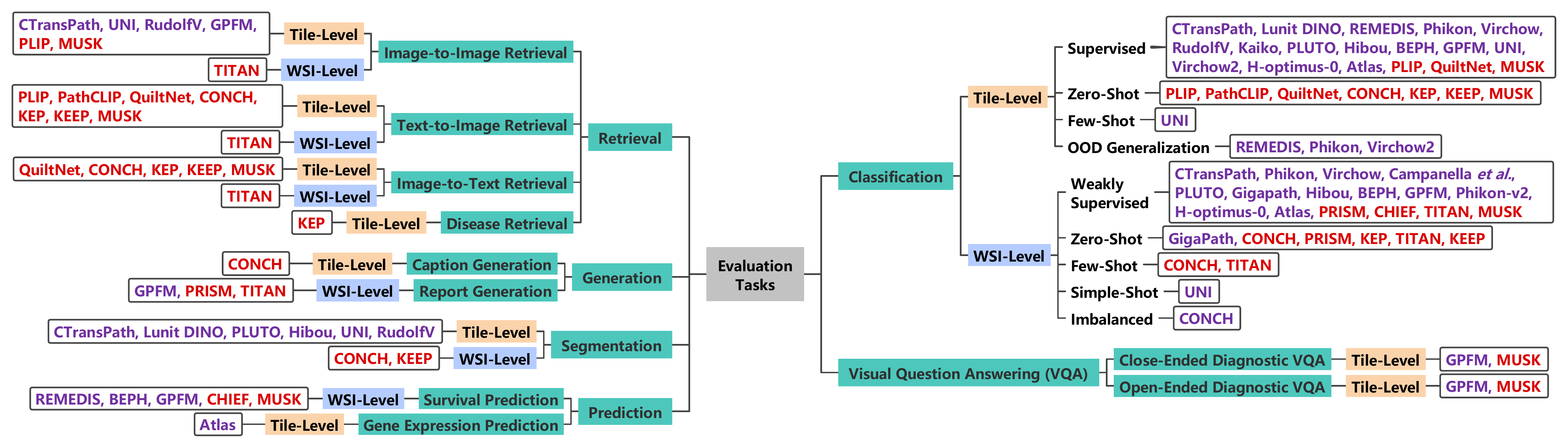}
    \vspace{-8mm}
    \caption{Taxonomy of evaluation tasks for pre-trained CPathFMs. Uni-modal and multi-modal CPathFMs are highlighted in \textcolor[rgb]{0.439, 0.188, 0.627}{\textbf{purple}} and \textcolor[rgb]{0.851,0,0}{\textbf{red}}, respectively.}
    \vspace{-5mm}
    \label{fig:evaluation-tasks}
\end{figure*}

\textbf{CoCa-based CPathFMs.}
The multi-modal decoder specifically designed for CoCa enhances the cross-modal capabilities of CPathFM, contributing to its application in the latest CPathFMs. CONCH \cite{CONCH} and PRISM \cite{PRISM} both pre-train an image encoder on pathology datasets using the iBOT and DINOv2 frameworks, respectively, and then further conduct joint visual-language pre-training within the CoCa framework. The difference is that PRISM extends the image encoder to the WSI-level using a Perceiver network \cite{Perceiver} and employs WSIs along with their corresponding clinical reports for training.
Additionally, MUSK~\cite{MUSK} first trains the image and text encoders separately on unpaired pathology images and text tokens using masked data modeling within the BEiT-3 framework. Then, it aligns the two modalities under the CoCa framework.
Building upon prior work, TITAN \cite{TITAN} develops a multi-modal whole-slide foundation model primarily designed for training a slide encoder. Its pre-training process is divided into three stages. First, a slide encoder is trained under the iBOT framework with positional encoding incorporated. Subsequently, under the CoCa framework, the slide encoder is trained at both the tile-level and WSI-level, facilitating the gradual development of the model's ability to comprehend and generate meaningful vision-language representations for WSIs.

In addition to the methods investigated in this work, there are other approaches that focus on multi-modal data beyond text and images, such as chest X-ray images~\cite{BiomedCLIP}, genetic sequences~\cite{TANGLE}, \textit{etc.,} which fall outside the scope of this survey focused on histopathology staining images.
Moreover, some works focused on developing multi-modal large language models (MLLMs) as generative foundation AI assistants for pathologists~\cite{PathCLIP} are not included, as these works typically align multi-modal CPathFMs mentioned above with existing LLMs. 
The emphasis of these models is on enhancing VQA capabilities rather than training a general feature extractor.

%% file: sections/tasks.tex


CPathFMs do not target a specific task during the pre-training phase. Instead, a wide range of evaluation tasks are employed after pre-training to assess the model's ability to extract features from pathology data. These tasks are diverse, and the evaluation tasks for each CPathFM are not standardized, making it challenging to establish a unified benchmark for CPathFMs. Therefore, we provide a summary of the evaluation tasks along with the CPathFMs performing them, as illustrated in Figure~\ref{fig:evaluation-tasks}.
We first categorized the evaluation tasks into six major perspectives based on their application objectives, followed by a further subdivision according to their specific objectives (e.g., focusing on tile-level or WSI-level). On this basis, we also considered variations in task settings (\textit{e.g.,} supervised or zero-shot learning). Finally, we summarized which CPathFMs were used to evaluate each type of task.


A series of complex pathology tasks, such as cancer subtyping, biomarker detection, and mutation prediction, are essentially classification problems. As the most commonly used evaluation task, classification has been widely studied in both tile-level and WSI-level under supervised, few-shot, and zero-shot settings. Unlike fully supervised learning, few-shot and zero-shot learning aim to perform on evaluation tasks when the pre-trained model has seen only a small portion or none of the training samples.
Classification tasks at the WSI-level are typically weakly supervised, meaning that such tasks only have global annotations (WSI-level) without details of internal regions. If a CPathFM only extracts tile-level features, a WSI-level representation through an aggregator network needs to be obtained when applying this task.
In addition to the three common settings mentioned above, some CPathFMs have also evaluated the model's out-of-distribution generalization ability under settings where there is a distribution shift between the training and testing datasets. Pathology images from different institutions, different staining methods, and different modalities can all contribute to distribution shifts. Additionally, CONCH classified rare diseases to validate its performance on imbalanced data.

Other task types include retrieval, generation, segmentation, prediction and VQA, which also involve both tile-level and WSI-level tasks. Among these, cross-modal retrieval, generation and VQA tasks challenge the cross-modal capabilities of CPathFMs.
In addition to qualitative analysis, some CPathFMs have undergone qualitative analysis. Virchow, RudolfV, BEPH and PLIP perform dimensionality reduction and clustering on the representations and observe the results.


 

%% file: sections/future_work.tex

\textbf{Trustworthy CPathFMs}
ensures fairness, explainability, security, and transparency. Fairness is especially crucial, as predicted outcomes should be independent of sensitive attributes, such as race, to avoid potential biases in clinical applications. Enhancing the explainability of CPathFMs is also essential to gaining the trust of pathologists and clinicians, as deep learning models often operate as black boxes. Furthermore, addressing security vulnerabilities in CPathFMs, such as adversarial attacks, is necessary to prevent manipulation of model predictions. Finally, transparency in model development, dataset curation, and evaluation procedures is crucial for reproducibility and regulatory approval, ensuring CPathFMs can be safely deployed in clinical workflows.

\textbf{Developing CPathFMs for MxIF Imaging.}
Unlike H\&E and IHC staining, MxIF captures spatial distributions of multiple biomarkers simultaneously, offering richer biological insights into the tumor microenvironment. However, training foundation models on MxIF images presents challenges, including higher dimensionality, complex signal processing, and the need for precise biomarker alignment. Future research should focus on building CPathFMs that can effectively extract meaningful representations from MxIF data while addressing these computational challenges.


\textbf{Standardized Benchmarking Datasets and Evaluation Metrics for CPathFMs.}
The current landscape lacks a uniform set of evaluation metrics that can systematically compare different models across a wide range of pathology tasks. A standardized benchmark dataset incorporating diverse tissue types, staining methods, and multi-institutional sources would significantly enhance model generalization and comparability. Additionally, defining clear evaluation indicators would allow the research community to assess the robustness, fairness, and clinical utility of CPathFMs more effectively.

%% file: sections/conclusion.tex
This survey provides a review of existing computational pathology foundation models, examining challenges in pre-training datasets, adaptation strategies, and evaluation tasks, while offering a comparative analysis of their strengths and limitations. Finally, we have identified key research gaps and proposed potential directions for future advancements.